# Block Motion Based Dynamic Texture Analysis: A Review

Akhlaqur Rahman[1] and Sumaira Tasnim[2]

[1]*Department of Electrical and Electronic Engineering Uttara University, Bangladesh*
[2]*Department of Science Engineering and Health, Central Queensland University, Australia*

**ABSTRACT:** *Dynamic texture refers to image sequences of non-rigid objects that exhibit some regularity in their movement. Videos of smoke, fire etc. fall under the category of dynamic texture. Researchers have investigated different ways to analyze dynamic textures since early nineties. Both appearance based (image intensities) and motion based approaches are investigated. Motion based approaches turn out to be more effective. A group of researchers have investigated ways to utilize the motion vectors readily available with the blocks in video codes like MGEG/H26X. In this paper we provide a review of the dynamic texture analysis methods using block motion. Research into dynamic texture analysis using block motion includes recognition, motion computation, segmentation, and synthesis. We provide a comprehensive review of these approaches.*

## 1. INTRODUCTION

A large class of non-rigid objects is commonly observed in a real world scenario that possesses movement patterns with certain form of regularities. The motion assembly by a water streams, flock of flying birds, waving flags, and fluttering leaves are some examples of dynamic textures. Contemporary literature coined the term *dynamic texture* (Figure 1) to refer to the movement pattern in the image sequences of these non-rigid objects [1].

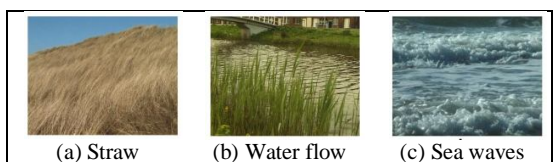

(a) Straw  (b) Water flow  (c) Sea waves
Figure 1: Dynamic textures [10]

The phenomena observed in dynamic textures have prompted many researchers in the computer vision community to formulate techniques to analyze these distinctive motion patterns. Appearance based approaches [2][3] aim to identify visual patterns (intensity based) based features to characterize them. As the uniqueness of the movement associates with the motion, appearance based approaches do not perform as good. Motion based approaches on the other hand, computes motion vectors first and computes features from the motion patterns to analyze dynamic textures. Two types of motion computation approaches are observed: (i) normal flow based [4][5] and (i) block motion based. Normal flow is a computationally inexpensive approach to approximate for the optical flow. Block motion is low resolution representation of motion filed readily available with the video codecs. As partial decoding of the video codecs can retrieve the motion vectors, use of them for the analysis will make the overall process suitable for real time applications like surveillance. Block motion is shown to perform as well as pixel level motion [6]. In this paper we concentrate on block motion based dynamic texture analysis methods.

The analysis tasks with dynamic texture using block motion ranges from recognition, separation, synthesis, segmentation, and retrieval. In the following sections we provide a comprehensive discussion on each of these methods using block motion.

## 2. DYNAMIC TEXTURE RECOGNITION

In [6]-[9] the authors have presented a Motion Co-occurrence Matrix (MCM) based approach to capture the motion distribution in the image sequences of dynamic textures. The images in the sequence are divided into equal sized blocks and Motion is computed for the blocks. Gray Level Co-occurrence Matrix (GLCM) is a commonly used method in image texture analysis. The authors in [6] have computed the co-occurrence matrices in two spatial directions and one temporal direction. Weights are calculated for each direction and combined using KL divergence. Recognition accuracy of 93% was achieved in [6] using *k*-NN classifier on DynTex data set [10]. In [9], 3D MCM was computed where the three indices of the matrix are *x* (spatial), *y*(spatial), and *t* (temporal). Recognition accuracy of 88% was reported on Szummer data set [11] using a decision tree classifier in [9] using block motion.

## 3. MOTION ESTIMATION FOR DYNAMIC TEXTURE

An accurate flow estimation technique independent of the type of dynamic texture is presented in [12]. This is achieved through the utilization of properties universal to all types of dynamic textures for motion estimation. Dynamic textures are motion patterns that possess spatiotemporal motion uniformity and in [12] the authors propose an algorithm that utilizes this uniformity cue for motion estimation to bring the different types of textures under a unified approach. Optical flow estimates obtained using our proposed algorithm are accurate and consistent with human observations. Dynamic textures are identified by their distinctive motion patterns and accuracy of optical flow estimates obtained using the method in [12] is also established by classifying a diverse set of dynamic textures with high accuracy. On an average 5.43% better recognition was achieved using the motion vectors in [12].

## 4. DYNAMIC TEXTURE SYNTHESIS

Texture synthesis, in general terms, refers to the process of generating a new texture that is similar to, but somewhat different from, the original texture. In order to generate a new temporal texture it is necessary to reproduce the visual quality as well as the dynamics of the original texture. A dynamic texture synthesis method is presented in [13][14]. During learning, texture movement is identified by computing motion vectors (computed by [12]) and then movement pattern within a texture is encoded by motion distribution statistics. Given a pair of seed image and corresponding motion frame, synthesis of a new texture is performed in two steps. The motion frames of the synthesized sequence are generated first from the seed motion frame using the a priori motion distribution statistics. The sequence of image frames is then constructed from the seed image frame, corresponding seed motion frame, and synthesized motion frames using guidelines of the motion vectors and local properties of dynamic textures. A diverse set of dynamic textures were synthesized by the method in [14]. Experimental results demonstrate the ability of the technique by producing visually promising dynamic textures.

## 5. FEATURE WEIGHTING AND RETRIEVAL METHODS FOR DYNAMIC TEXTURE

Feature weighing methods are commonly used to find the relative significance among a set of features that are effectively used by the retrieval methods to search image sequences efficiently from large databases. Block motion based approaches mostly use Motion Co-occurrence Matrix (MCM) features that are called abstract features [15][16]. The conventional feature weighting approach considers the features altogether and uses a heuristic search for large feature sets, as an exhaustive search is highly time consuming under such a scenario. Considering domain information, however, MCMs can be organized into a hierarchy and weights can be computed at each level in the hierarchy. Such an approach facilitates using an exhaustive search at levels with a smaller number of features, and thus improving the overall classification accuracy. In [15] a hierarchical approach for feature weighting is presented that provides a mechanism to control motion measure specific MCM, clique and domain weights at different levels. The conventional feature weighting approach is unable to control MCM weights domain wise and leaves room for some classification accuracy improvement. 96.03% recognition accuracy was obtained using the approach in [15]. A set of retrieval methods using MCM features is presented in [16][17][18].

## 6. SEGMENTATION OF DYNAMIC TEXTURES

The problem of detecting multiple dynamic textures by segmentation using block motion based MCM is presented in [19][20]. MCM features are computed from motion distribution of the segments and compared with a set of ground truth unique textures to find the class of the segmented textures. Accuracy of detection by segmentation methods is in general high, as even a gross segment can accurately detect the underlying single texture. This motivated the authors in [19][20] to use block motion. As argued by the authors, are in the literature for long, straightforward extension of conventional image texture segmentation techniques to dynamic texture will be ineffective, as images have no time dimension, whereas significant information is carried by this dimension to characterize a dynamic texture. The authors thus presented a spatiotemporal segmentation method using MCM as the underlying motion distribution statistics. Segmentation results by the Spatiotemporal Split and Merge segmentation based detection method differentiated multiple dynamic textures with accurate number of segments.

## 7. SEPARATION OF MULTIPLE DYNAMIC TEXTURES

In [21][22][23] the authors presented a two phase texture detection process. In the faster feature based detection phase, only the presence of multiple textures is detected using MCM feature with a multi–texture dataset where the single texture image sequences are mixed with each other at different image volume ratio. Any target texture–mix detection is then confirmed by the slower segmentation based detection phase to eliminate false detections or isolate independent texture regions for true detections. In a continuous monitoring application, if the false detection rate is low, the frequency of the slower phase is negligible to that of the faster phase, hence the process can comply with the real time constraint.

For the faster detection phase, a mapping between the feature space of unique dynamic textures and that of a sequence where multiple dynamic textures are present, so that the feature space of individual and multiple temporal textures can be compared directly to find which temporal textures are present. More precisely, A linear correspondence between MCMs of individual temporal textures and that of their mixture in an image sequence were exploited and feature based multiple temporal texture detection method was developed. The correctness of the method is both analytically and empirically established. A set of recognition experiments, while conducted on a diverse set of temporal texture mixtures, reveals that the method achieves detection accuracy as high as 92.5%.

## 8. CONCLUSIONS

In this paper we have presented a review of a group of dynamic texture analysis methods that use block motion as underlying feature. A wide range of analysis methods including recognition, synthesis, segmentation, separation etc. are covered. In future we aim to investigate new approaches to use block motion based features for novel analysis applications.